\documentclass{template/svproc}
\usepackage{url}
\usepackage{graphicx}
\usepackage{amsmath}

\begin{document}
\mainmatter              
\title{Sim2Plan: Robot Motion Planning via Message Passing between Simulation and Reality}
%
\authorrunning{Published as a conference paper at FTC 2023}
\titlerunning{Sim2Plan}  
%
\author{
Yizhou Zhao\inst{1} \and Yuanhong Zeng\inst{1} \and Qian Long \inst{1}
\and \\Ying Nian Wu\inst{1}
\and Song-Chun Zhu\inst{1}}
\institute{Department of Statistics, University of California, Los Angeles\\
\email{yizhouzhao@g.ucla.edu}\\ 
}

\maketitle              

\begin{abstract}
Simulation-to-real is the task of training and developing machine learning models and deploying them in real settings with minimal additional training. This approach is becoming increasingly popular in fields such as robotics. However, there is often a gap between the simulated environment and the real world, and machine learning models trained in simulation may not perform as well in the real world. We propose a framework that utilizes a message-passing pipeline to minimize the information gap between simulation and reality. The message-passing pipeline is comprised of three modules: scene understanding, robot planning, and performance validation. First, the scene understanding module aims to match the scene layout between the real environment set-up and its digital twin. Then, the robot planning module solves a robotic task through trial and error in the simulation. Finally, the performance validation module varies the planning results by constantly checking the status difference of the robot and object status between the real set-up and the simulation. In the experiment, we perform a case study that requires a robot to make a cup of coffee. Results show that the robot is able to complete the task under our framework successfully. The robot follows the steps programmed into its system and utilizes its actuators to interact with the coffee machine and other tools required for the task. A noteworthy observation from the experiment is the speed and accuracy with which the robot completed the task. The robot can make a cup of coffee relatively quickly compared with traditional robot planning and control methods, and its movements were precise and efficient. Overall, the results of this case study demonstrate the potential benefits of our method that drive robots for tasks that require precision and efficiency. Further research in this area could lead to the development of even more versatile and adaptable robots, opening up new possibilities for automation in various industries.

\end{abstract}
\newpage
\section{Introduction}

\begin{figure}[t]
    \centering
    \includegraphics[width = 0.75\textwidth]{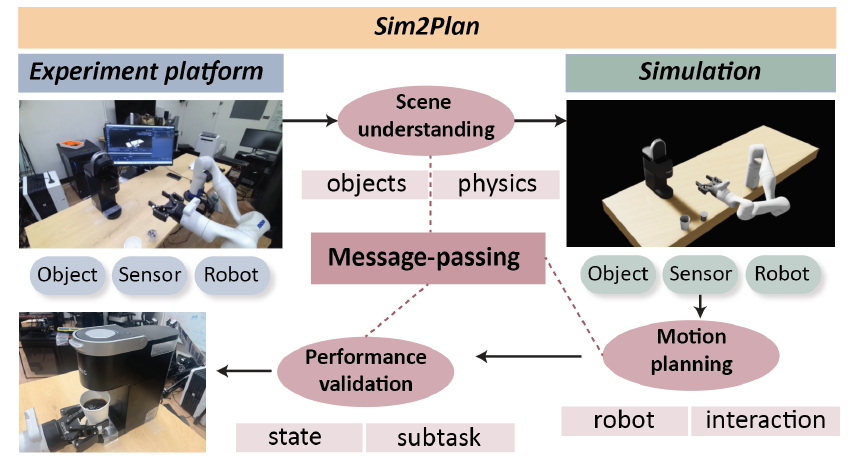}
    \caption{An overview of our \textsc{Sim2Plan} framework.A high-level schematic of the three main components that make up the Sim2Plan framework: the experiment platform, the simulation engine, and the message-passing pipeline.}
    \label{fig:framework}
\end{figure}

Traditionally, training a robot in a real setting involves designing the task, examining and setting up hardware, programming the robot, and testing the performance~\cite{ibarz2021train}. These steps require careful planning, design, and execution, as well as ongoing evaluation and refinement. In general, training a robot in the real world can be expensive in terms of cost and time, especially when optimal performance and safety need to be ensured.

Recently, Simulation-to-reality (Sim2Real) has been active in robotics. Driven by advances in physics-based simulation, machine learning, and AI-based benchmarking, Sim2Real techniques make it more efficient and accessible for robot training and application in the real world. Compared with traditional robot training methods, Sim2Real can be less expensive than traditional robot training methods because they do not require as many environmental resources~\cite{hofer2021Sim2Real}. It can also increase safety, enable faster iteration, allow more precise and complex control, and improve the robot's performance from more training data.

Although Sim2Real techniques offer many advantages for robot training and deployment, they also introduce some challenges related to the Sim2Real gap. For example, simulated environments may not perfectly match the real world: simulation can be hard to perfectly replicate the complexities and nuances of the real world~\cite{hofer2021Sim2Real}. Besides, robots trained in simulation may struggle to generalize their learned behaviors and policies to new, unseen environments in the real world, which limits the robot's performance and adaptability in diverse environments~\cite{tan2018sim}. In addition, real-world environments are inherently uncertain and variable, with unpredictable factors such as lighting, background, and human interactions~\cite{ibarz2021train}.

To address the challenges introduced by the Sim2Real gap, we introduce \textsc{Sim2Plan}, a framework that leverages the strengths of simulation for robot motion planning in real-world deployment. \textsc{Sim2Plan} combines advanced simulation techniques with transfer learning and domain adaptation to enable robots to learn from simulated environments and transfer those learned behaviors to the real world. 

Specifically, \textsc{Sim2Plan} consists of three main components: a real world experiment platform, a simulated environment, and a message-passing pipeline. The core idea behind \textsc{Sim2Plan} is to use the message-passing pipeline to interchange the information between the real-world experiment and the simulation while constantly checking and correcting the information gap. The message-passing pipeline (composed by \textit{scene understanding}, \textit{robot motion planning}, and \textit{performance validation}) is targeted to match the scenes, robots, and experiments between reality and simulation. The scene understanding module collects data from real experiments and constructs a digital twin in simulation. It also provides essential information for the robot motion planning module, which generates actions for the robot both in simulation and in the physical environment. Lastly, the performance validation Module evaluates the robot's actual behavior against its simulated counterpart, allowing for continuous improvement and optimization of the system. Overall, our framework enables efficient and accurate robot control through seamless integration between virtual and real-world environments.

We evaluate \textsc{Sim2Plan} on a coffee-making case study. This study requires the robot to perform object manipulation tasks through motion planning.  \textsc{Sim2Plan} demonstrate its effectiveness in improving the performance and generalization of robotic systems. Our results show that \textsc{Sim2Plan} can significantly reduce the amount of real-world training time required while enabling robots to perform effectively in diverse and challenging randomized environments.

Overall, \textsc{Sim2Plan} represents a promising approach to addressing the challenges introduced by the Sim2Real gap, and has the potential to significantly improve the efficiency and effectiveness of robot training and deployment in the real world.

\section{Related Work}
A considerable body of research is related to our study, including work on simulating real-world environments for robots, motion planning algorithms, and embodied AI simulation.

\subsection{Sim2Real}

Recent advances in simulation-to-reality (Sim2Real) transfer have enabled robots to learn complex manipulation tasks in simulation and apply these skills to the real world. For instance, researchers have used Sim2Real transfer to teach robots to grasp objects with greater accuracy~\cite{horvath2022sim2real}, navigate through challenging environments~\cite{truong2021bi}, and even perform tasks like pouring liquid into a cup~\cite{gong2023arnold}. These advances are made possible by using machine learning algorithms to train robots in simulation and then fine-tuning the learned skills in the real world. Additionally, advancements in hardware technology, such as high-fidelity simulators and robust robotic systems~\cite{makoviychuk2021isaac} , have contributed to the success of Sim2Real transfer. As a result, Sim2Real transfer is becoming an increasingly popular approach for developing more capable and versatile robots that can operate effectively in dynamic, real-world environments.

\subsection{Robot Motion Planning}

Recent studies have focused on developing more advanced algorithms for robot motion planning that can handle increasingly complex scenarios. Some of these approaches involve using machine learning techniques to generate plans based on past experience~\cite{morales2005machine}, while others leverage cloud computing resources to distribute computationally intensive tasks among multiple servers~\cite{vick2015robot}. Other areas of interest include improving plan robustness to uncertainty and ensuring safe interactions with humans in shared workspaces~\cite{gao2020joint}. Ultimately, the goal of these efforts is to enable robots to perform tasks autonomously and efficiently in dynamic and uncertain environments~\cite{latombe2012robot}.

\subsection{Embodied AI Simulators}

Embodied AI is intelligence that emerges through interacting with environments \cite{franklin1997autonomous}. The growing interest in embodied AI fosters the development of embodied AI simulators, which serve as benchmarks~\cite{das2018embodied} to train and develop intelligent systems before deploying them in the real world. The simulators typically address three typical AI research tasks: visual exploration, visual navigation, and embodied question-answering~\cite{duan2022survey}. In visual exploration, the agent navigates through the environment, processes visual information, identifies objects, and learns their spatial relationships~\cite{jia2022learning}. In visual navigation, the agent knows to plan its route, avoid obstacles, and adapt its strategy based on environmental changes~\cite{zhao2021luminous}. Finally, embodied QA tasks involve AI agents answering questions or reasoning about their environment based on their egocentric perceptions.




\section{Framework}

In this section, we will discuss the various components that make up our \textsc{Sim2Plan} framework, including establishing an experimental platform in the real world, creating a simulated environment, and implementing a robust messaging-passing pipeline. We show these in Figure~\ref{fig:framework}.

The experiment platform serves as the interface for the real-world robot environment (Section~\ref{sec:experiment_plaform}). The simulation part acts as the digital twin of the experiment platform (Section~\ref{sec:simulation}), allowing for accurate modeling and prediction of system behavior. Finally, the core element of the framework is the message-passing pipeline~(Section \ref{sec:experiment}), which facilitates seamless communication between the experiment platform and the simulation engine.

\subsection{Experiment Platform}
\label{sec:experiment_plaform}
The \textsc{Sim2Plan} framework requires the creation of a physical experimentation platform in which the simulated models can be tested in the real world. \\

\noindent\textbf{Robot}.
Setting up a robot in a physics space requires careful consideration of several factors, such as the size and shape of the workspace, the type of tasks the robot needs to perform, the sensors required to perceive its surroundings, and the actuators necessary for motion control.

To set up the robot for our experiment, we followed several steps. Firstly, we designed a fixed area as our workspace, where the robot would perform its tasks. This was necessary to ensure that the robot would operate within a defined and controlled environment, which would help us to measure its performance accurately.

Next, we chose the robot arm as our primary training target. The robot arm is a crucial component that enables the robot to manipulate objects and perform tasks in the environment. By training the arm, we could help the robot develop the skills needed to perform its functions effectively.

Finally, we select the gripper as the end-effector for the robot. The gripper is a device that allows the robot to grasp and manipulate objects, while the end-effector is the component attached to the end of the robot arm and is responsible for performing specific tasks, such as picking up and moving objects. By carefully selecting the gripper and end-effector, we could ensure that the robot has the necessary tools to perform its tasks effectively and efficiently.\\

\noindent\textbf{Sensor}.
We use a single RGB camera without a depth sensor as the sole sensor in the scene for the following reasons:

\textit{Simplicity}: A monocular camera setup is often the most straightforward option, requiring fewer resources and less complex calibration than stereo or multi-camera systems. This makes it suitable for smaller projects or prototyping purposes where complexity may not be desirable.
    
\textit{Portability}: By relying exclusively on an RGB camera, the system becomes highly portable since no additional sensors need to be integrated into the setup. This allows for quick deployment across multiple platforms or environments without significant modifications.
    
\textit{Versatility}: Despite being a basic configuration, an RGB camera can still capture valuable information for various perception tasks, including object detection, segmentation, and even Simultaneous Localization And Mapping (SLAM). These algorithms rely heavily on visual cues from images, making the RGB camera a sufficient input data source.

While other configurations might offer greater robustness or accuracy, an RGB camera remains a practical choice due to its ease of implementation, affordability, and broad applicability. As technology advances, these benefits continue to make it a viable option for many real-world scenarios.\\

\noindent\textbf{Object}.
When considering the interaction between the robot's tool and the objects in the scene, we must account for their physical properties. Our framework will focus on three distinct categories: rigid bodies, soft bodies, and fluids. Each type presents unique challenges when attempting to manipulate or interact with the environment.

\textit{Rigid bodies}: objects made up entirely of solid material, such as metal or plastic, are classified as rigid bodies. They maintain their shape under external forces and not deform unless subjected to extreme stress or impact. Manipulating rigid bodies requires careful consideration of their mass distribution, center of gravity, and friction coefficients. Tools designed for rigid bodies typically have high stiffness and low compliance to minimize deflection and ensure stable interactions. Examples of rigid bodies encountered in our scenario could include cups, coffee machines, or furniture pieces.

\textit{Soft bodies}: Unlike rigid bodies, soft bodies exhibit some degree of elasticity or compressibility. Soft materials, such as foam, rubber, or fabric, behave differently than hard solids when subjected to external loads. Interacting with soft bodies demands special attention to contact mechanics, deformation modeling, and damping effects. 

\textit{Fluids}: Fluid would introduce new complexities into the equation due to its continuous nature and nonlinear behavior. Flow patterns, turbulence, and viscosity variations play crucial roles in understanding how fluids respond to pressure, temperature, or velocity changes. In addition, robots operating within fluid environments need to cope with issues related to buoyancy, drag, and other dynamic properties. 

\subsection{Simulation}
\label{sec:simulation}
In this section, we will describe the setup of the digital twin of the experiment in the simulation environment, which involves retaining simulated robot, sensor, and objects.\\

\noindent\textbf{Simulation Engine}.
To create a comprehensive simulated training program for robots, we choose \textsc{Nvidia Omniverse}~\cite{omniverse} as the development platform for our \textsc{Sim2Plan}. \textsc{Omniverse} boasts cutting-edge simulation features that enable efficient and dependable representation of rigid bodies, soft bodies, articulated objects, and fluids. Furthermore, the platform supports seamless integration of Python scripts, enabling access to a vast array of open-source and third-party libraries. Another advantage of using \textsc{Omniverse} lies in its advanced ray tracing technology, allowing for breathtakingly realistic renderings.\\

\begin{figure}
    \centering
    \includegraphics[width = 0.75\textwidth]{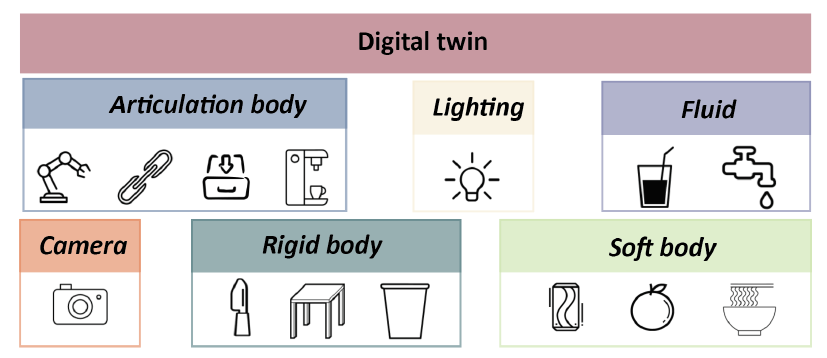}
    \caption{ Digital twin components. An overview of the key elements involved in creating a comprehensive digital twin, including rigid bodies, lighting, fluids, cameras, articulation bodies, and soft bodies.}
    \label{fig:digital_twin}
\end{figure}

\noindent\textbf{Digital Twin}.
A digital twin is a virtual replica of a physical entity created through computer simulations and sensor data~\cite{tao2018digital}. The purpose of establishing a digital twin is to provide a dynamic, interactive representation of the original object, allowing for better analysis, prediction, and optimization of its performance. First, we meticulously scrutinize the surroundings and concentrate on configuring the digital twin for the robot, camera, and task items (see Figure~\ref{fig:digital_twin}). Afterward, we utilize the digital twin to reenact diverse situations and gauge the system's execution for the experiment.

\section{Message-Passing Pipeline: An Experiment}
\label{sec:experiment}
In this section, we test our \textsc{Sim2Plan} framework in a case study: make coffee by a coffee machine. Then, we introduce how we set up the digital twin, discuss implementing the message-passing pipeline, and demonstrate the results.

\subsection{Preparation}
To create a coffee-making experiment's digital twin, we first gather information about its physical properties, such as dimensions, materials used, and internal components. Then, we use computer-aided design (CAD) software to model the coffee machine digitally. Next, we simulate the behavior of the coffee machine using physics engines in \textsc{Omniverse}. These simulations consider gravity, friction, and other forces acting on the device during operation.

\begin{figure}[ht]
    \centering
    \includegraphics[width=0.95\textwidth]{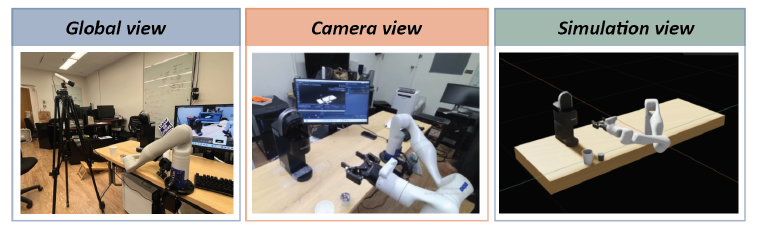}
    \caption{Different views of the experiment. The global view shows the overall setting of the experiment. The camera view shows what can be seen from the camera. And the simulation view shows the digital twin of the experiment.}
    \label{fig:view}
\end{figure}

The experiment is captured from multiple angles to give a comprehensive setup overview (see Figure~\ref{fig:view}). The global view provides a broad perspective of the experimental arrangement, showcasing the interaction between the physical objects and the virtual representation. Meanwhile, the camera view offers a closer look at specific aspects of the experiment, highlighting details that would be applied for object detection. Lastly, the simulation view displays the digital twin of the investigation, allowing us to visualize the system's inner workings and plan the robot's behavior. Together, these views offer comprehensive insights into the experiment and enable more informed decision-making.

\subsection{Scene understanding}

\textbf{Prior knowledge.}
We first utilize prior knowledge to gain a basic understanding of the scene layout. Prior knowledge refers to the fixed measures within the scene, such as the sizes~$\hat{s}_i$ of objects like the table, robot, coffee machine, cup, and coffee capsule. Since the camera, table, and robot positions remain unchanged throughout the experiment, we also measure their respective positions~$\hat{p}_i$. By incorporating these measurements into our analysis, we enhance our ability to interpret the visual input from the camera view and obtain a clearer picture of the scene.
\begin{align}
\hat{S}_{\text{prior}} &= \{\hat{s}_{\text{table}},\hat{s}_{\text{robot}}, \hat{s}_{\text{coffee\_machine}},\hat{s}_{\text{cup}},\hat{s}_{\text{capsule}}\} \\
\hat{P}_{\text{prior}} &=\{\hat{p}_{\text{table}},\hat{p}_{\text{camera}}, \hat{p}_{\text{robot}}\}
\end{align}

\noindent\textbf{Object detection.}
To better understand the scene layout, we employ the object detection module to identify objects present in the scene. Two state-of-the-art deep learning-based algorithms are used for this purpose: Open-Vocabulary Object Detection (OWL-ViT)~\cite{minderer2022simple} and Grounding DINO~\cite{liu2023grounding}. Both methods take the image and text prompt as the inputs and leverage powerful vision transformers to detect and localize objects within the image frame, providing accurate bounding boxes and class labels for each detected instance. This information serves as crucial contextual awareness for subsequent tasks involving manipulation planning and execution. \\

\noindent \textbf{Vision inference.}
Besides using the object detection module, we also gather relevant metadata about the camera to ensure optimal calibration and accuracy. Precisely, we determine the camera's focal length $(f_x, f_y)$, resolution, and principal point $(c_x, c_y)$, which are essential parameters for correcting lens distortion and projecting 3D points onto the 2D image plane. With all this information, we have a solid foundation for building a reliable and effective perception system tailored to our needs. 

\begin{figure} [h]
    \centering
    \includegraphics[width=\textwidth]{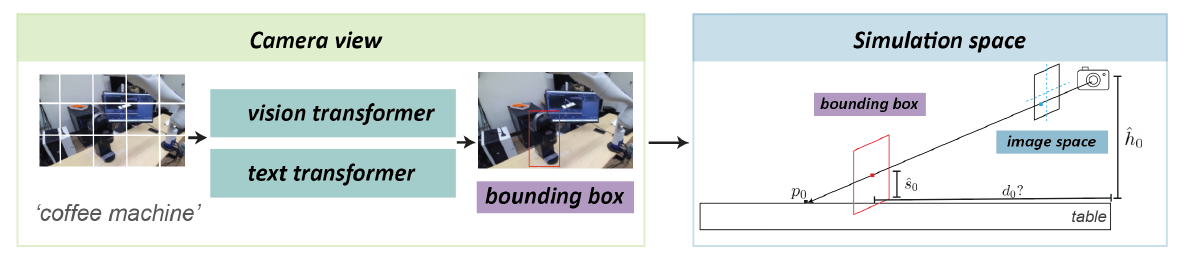}
    \caption{Obtaining object 3D position from 2D camera view. A visual explanation of the process used to estimate the 3D position of an object from a single 2D image captured by a camera. After the getting the bounding box (thus the object center) in the image space, we can project the point $p_0$ to the ground (table) in 3D space. Since the size of the object $\hat{s}_0$ and the position of the camera $\hat{h}_0$ are known as prior knowledge, we can thus determine the the object's 3D position $d_0$.}
    \label{fig:projection}
\end{figure}

Figure~\ref{fig:projection} demonstrates the steps involved in estimating the 3D position of an object from a 2D camera view. The process begins with a computer vision model (left side), which uses the input image to predict the location of the object. Once the bounding box in the image space of the object is obtained, the next step (right side) utilizes geometric principles to calculate the 3D position of the object relative to the camera's field of view. Specifically, this requires knowledge of the camera's intrinsic parameters (e.g., focal length, principal point) and extrinsic parameters (e.g., rotation matrix, translation vector). These values allow us to project the bounding box onto the 3D world coordinate frame, resulting in the final estimated position of the object in 3D space. 

Compared to Owl-Vit, the Ground-DINO model performs better in detecting objects such as cups, coffee machines, and coffee capsules. Leveraging vision inference techniques, the final prediction error for the 3D position of these objects can be controlled within $5$ mm, enabling precise robot motion planning. This level of accuracy is sufficient for our real-world experiment.

\subsection{Robot Motion Planning}
After obtaining the scene information from the prior knowledge and the vision module, we use the Riemannian Motion Policy (RMP)~\cite{ratliff2018riemannian} as the motion policy controller for the robot in its digital twin simulation. RMP has the following advantages. Firstly, RMP considers the geometry of the configuration space (c-space), which allows for smooth and efficient trajectories even when working close to singularities or other nonlinear regions. Secondly, RMP ensures that the resulting motions satisfy constraints on joint velocities, accelerations, and torques, making it suitable for robots with limited dynamic capabilities. Thirdly, RMP enables real-time optimization of motion plans based on sensor feedback, enabling adaptive behaviors that respond to environmental changes. Finally, RMP simplifies the design of complex motion sequences, reducing the computational burden required for generating feasible solutions.

After applying the RMP as the motion policy controller, we generate collision-free paths for the robot using Rapidly-exploring Random Tree (RRT)~\cite{lavalle2001rapidly}. The RRT algorithm constructs a tree data structure that grows randomly in the high-dimensional configuration space until it reaches a solution. New nodes are sampled randomly at each iteration and connected to existing ones if they lie within a certain distance threshold. We then check whether newly added nodes violate collision constraints with environmental obstacles. If so, we reject them; otherwise, we add them to the tree. Once the tree spans the entire configuration space, we extract a valid path between the initial and final configurations. Our approach leverages the strengths of both RMP and RRT, allowing us to achieve safe and efficient motion planning under uncertainty.

\begin{figure} [t!]
    \centering
    \includegraphics[width=\textwidth]{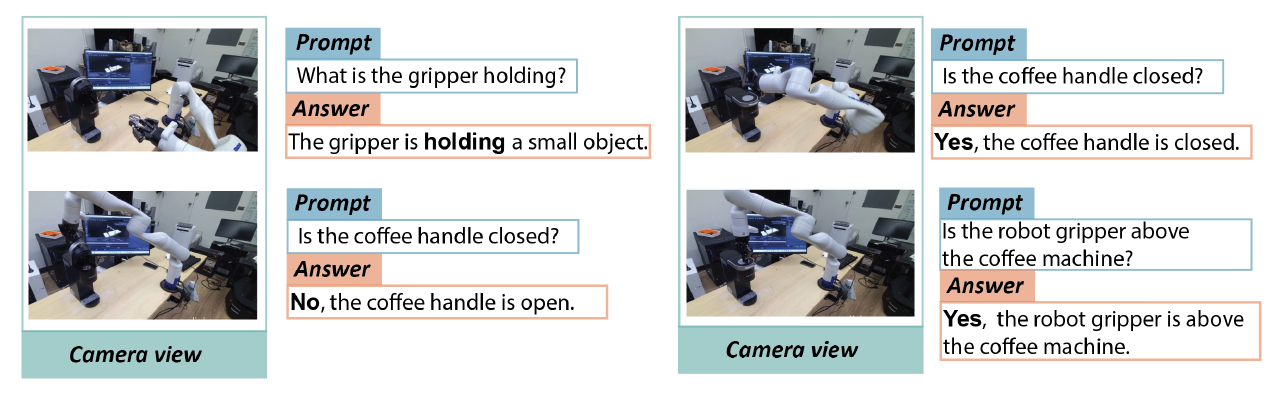}
    \caption{Task completion verification by MiniGPT-4: the image from the camera view and the prompt are input, and we check the keyword in the response (answer) to verify the task is complete.}
    \label{fig:gpt4}
\end{figure}

\subsection{Performance Validation}
We assess the robot's performance by verifying its configuration and task completion against simulations. By comparing the actual robot configuration with the planned one, we ensure that the physical robot adheres to the desired path generated during simulation. Additionally, we validate the successful completion of subtasks by evaluating the task configuration in the real environment. These checks enable us to confirm that the robot operates correctly and achieves its intended goals, thereby improving overall reliability and effectiveness. \\

\noindent \textbf{Robot Configuration}.
We continually compare the actual joint states $j_i$ (and gripper state $g_i$) with those predicted by the simulation. By doing so, we can ensure that the physical robot follows the intended instructions and performs according to expectations. The deviations or errors identified during this process can be addressed and corrected. If a large deviation is detected, we immediately stop the task execution to prevent the failure case from causing any safety concerns. Through this validation step, we can  improve the reliability, effectiveness, and safety of the robotic system. \\

\noindent \textbf{Task Completion}.
In addition, we continually compare the robot's performance in the real world with simulation by verifying the completion of subtasks. To ensure that each subtask, such as \textit{pick up the cup} or \textit{place the cup}, is executed correctly, we apply visual question answering (VQA) techniques~\cite{antol2015vqa} to verify the goal conditions from the camera's perspective. Specifically, we employ the newly released MiniGPT-4~\cite{zhu2023minigpt} module to perform the VQA task in practice. This allows us to accurately assess the robot's ability to accomplish specific subtasks and identify discrepancies between the simulated and real-world environments from vision-based prediction. 

Figure~\ref{fig:gpt4} illustrates verifying subtask completion using Visual Question Answering (VQA). To do this, we first compare the current robot configuration with the planned one after the planning and execution stages of a subtask. Next, we employ the MiniGPT-4 model to analyze the real-time visual input from the camera. Then, based on the specific context of each subtask, we formulate appropriate prompts for the VQA module and receive the corresponding answers from MiniGPT-4. Subsequently, we examine the keywords in these responses (such as \textit{yes} or \textit{no}) to confirm the successful completion of the subtask. This method provides an accurate and timely evaluation of the robot's progress, ensuring that it meets the requirements of each step along the way.

\begin{figure}[t]
    \centering
    \includegraphics[width=\textwidth]{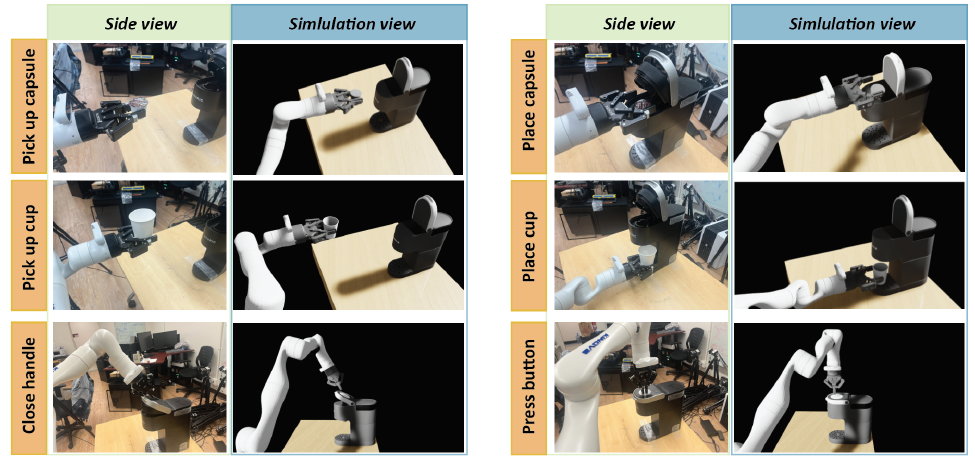}
    \caption{Real-world side camera view vs. simulated screenshot for subtask examples. Comparison between real-world images captured through a side camera and their simulated counterparts in the virtual environment.}
    \label{fig:examples}
\end{figure}

\subsection{Results}
Our experiments have demonstrated the effectiveness of our \textsc{Sim2Plan} framework for zero-shot robot motion planning (without training the robot in real space). In the first set of trials, where the positions of the coffee machine, coffee capsule, and cup are fixed, our framework achieved a remarkable 90\% success rate out of 20 attempts. When the positions of the capsule and cup are randomized, our framework could still successfully pick up the items 83.3\% and 76.6\% of the time, respectively. Despite the increased difficulty due to randomization, our framework managed to complete 75\% of a total of 20 trials. Overall, these results highlight the robustness and adaptability of our \textsc{Sim2Plan} framework in various environments and situations, making it a promising tool for robotic manipulation tasks.

Figure~\ref{fig:examples} presents several examples of subtasks performed by a robot from a side camera view in the real world and their corresponding screenshots taken from the simulation environment. Each block displays two images side by side, with the left one being the real-world snapshot and the right one showing the simulated scenario. These comparisons showcase how closely the simulation matches the real-world environment, demonstrating the validity of our proposed framework for zero-shot robot motion planning.

Furthermore, our proposed method also has practical benefits when applied to real-world settings. Since the robot was trained in a zero-shot manner during testing, it did not require any additional fine-tuning or retraining after being deployed to new environments. This means that the robot could quickly adapt to new situations without incurring significant delays or costs associated with retraining. By leveraging the motion planning from the digital twin, our method effectively reduces the amount of time required to train robots for specific tasks, making them more versatile and useful in a wide range of industries.

\section{Conclusion}

Sim2Real is an increasingly popular approach in robotics that involves training and developing machine learning models in a simulated environment and deploying them in the real world with minimal additional training. However, there is often a gap between the simulated environment and the real world, limiting the effectiveness of machine learning models trained in simulation. 

In this work, we present \textsc{Sim2Plan}, a framework that utilizes a multi-stage message-passing pipeline that has been proposed to minimize the information gap between simulation and reality. \textsc{Sim2Plan} includes modules such as scene understanding, robot planning, and performance validation, and it is demonstrated through a case study involving a robot making a cup of coffee. The results of the experiment suggest that the proposed framework has the potential to drive robots for tasks that require precision and efficiency and could lead to the development of more versatile and adaptable robots that could increase productivity in various industries.

\bibliography{article} 
\bibliographystyle{abbrv}

\end{document}